%% file: sample-sigconf.tex
\documentclass[sigconf]{acmart}

\usepackage{booktabs} 
\usepackage{array}

\usepackage{graphicx}
\usepackage{wrapfig}

\usepackage{multirow}
\usepackage{mathtools}
\DeclareMathOperator*{\argmax}{argmax}
\usepackage{soul}

\usepackage{booktabs} 
\usepackage{graphicx} 
\usepackage{float}    

\usepackage{footnote}
\newcommand{\sysname}{DiffuBias}

\setcopyright{acmcopyright}

\acmDOI{xx.xxx/xxx_x}
\acmSubmissionID{102}
\acmISBN{979-8-4007-0629-5/25/03}

\acmConference[ACM]{Conference}
\acmYear{2025}
\copyrightyear{2025}

\acmArticle{4}
\acmPrice{15.00}

\begin{document}
\title{Debiasing Classifiers by Amplifying Bias with Latent Diffusion and Large Language Models} 

  
\renewcommand{\shorttitle}{SIG Proceedings Paper in LaTeX Format}

\author{Donggeun Ko}
\affiliation{%
  \institution{AiM Future}
  \country{South Korea}
}
\email{sean.ko@aimfuture.ai}

\author{Dongjun Lee}
\affiliation{%
  \institution{Maum AI}
  \country{South Korea}
}
\email{akm5825@maum.ai}

\author{Namjun Park}
\affiliation{%
  \institution{Sungkyunkwan University}
  \country{South Korea}
  }
\email{951011jun@skku.edu}

\author{Wonkyeong Shum}
\affiliation{%
  \institution{Sungkyunkwan University}
  \country{South Korea}
  }
\email{23stella@skku.edu}

\author{Jaekwang Kim}
\authornote{Corresponding Author}
\affiliation{%
  \institution{Sungkyunkwan University}
  \country{South Korea}
  }
\email{linux@skku.edu}


\input{main/01_abstract}

\maketitle

\input{main/02_introduction}
\input{main/03_related_work}
\input{main/04_methodology}
\input{main/05_experiments}

\input{main/06_discussion}
\input{main/07_conclusion}

%
%




\clearpage
\bibliographystyle{ACM-Reference-Format}
\bibliography{sample-bibliography}

\clearpage
\input{main/appendix}

\end{document}

%% file: main/01_abstract.tex
\begin{abstract}
 Neural networks struggle with image classification when biases are learned and misleads correlations, affecting their generalization and performance. Previous methods require attribute labels (e.g. background, color) or utilizes Generative Adversarial Networks (GANs) to mitigate biases. We introduce \sysname, a novel pipeline for text-to-image generation that enhances classifier robustness by generating bias-conflict samples, without requiring training during the generation phase. Utilizing pretrained diffusion and image captioning models, \sysname~generates images that challenge the biases of classifiers, using the top-$K$ losses from a biased classifier ($f_B$) to create more representative data samples. This method not only debiases effectively but also boosts classifier generalization capabilities. To the best of our knowledge, \sysname~is the first approach leveraging a stable diffusion model to generate bias-conflict samples in debiasing tasks. Our comprehensive experimental evaluations demonstrate that~\sysname~achieves state-of-the-art performance on benchmark datasets. We also conduct a comparative analysis of various generative models in terms of carbon emissions and energy consumption to highlight the significance of computational efficiency.   
\end{abstract}

%% file: main/02_introduction.tex
\section{Introduction}

Deep learning models have demonstrated outstanding performances across various computer vision tasks including image classification~\cite{geirhos2018imagenet}, object detection and image generation~\cite{goodfellow2020generative}. However, generalization and robustness of the models have often been one of the major issues when we utilize our trained model in real-world scenarios. As benchmark datasets do not fully reflect real-world scenarios to train deep neural networks, models are often prone to biases that exists in the dataset. Unintended bias may exhibit high spurious correlations with the target label causing the model to learn these features during training. Thus, the model fails to generalize over the training distribution.

In the context of deep learning frameworks at large, bias refers to the tendency of these models to disproportionately weigh specific attributes or patterns—such as the shapes of an object, colors and textures within images, or their color—during the process of representation learning and subsequent classification task. Unintended biases include colors~\cite{kim2019learning}, textures, background, and gender~\cite{liu2015deep_celeb}. These unintended biases can be further categorized into \textit{task-relevant} and \textit{task-irrelevant} features where they may exist in a dataset. To clarify these two concepts, we can imagine an image of a horse in two different backgrounds. One image of a horse has a grass field in a plain as a background, while the other image of a horse is a racing horse in a rough field. In these images, the background of the location of the horse is a \textit{task-irrelevant} feature where these features mislead the model to learn unrelated features for the given task (i.e., image classification). In addition, if an image of a horse in a grass field is extensively abundant than the horse in a rough field, we can say that the grass field is \textit{bias-align} feature while the rough field is a \textit{bias-conflict} feature.

Datasets encountered in the wild, or 'real-world' datasets, are predominantly imbalanced and exhibit bias. Consequently, models trained on benchmark datasets, which possess a different bias distribution, often underperform on real-world datasets. The variations in classifier efficacy, as highlighted in Table~\ref{tab:vanilla_unbiased}, illustrate the pivotal issue of bias within datasets. The presence of biases significantly hampers model performance, with accuracy seldom exceeding $50\%$ on numerous datasets, including CCIFAR-10, BAR, and Dogs \& Cats. Notably, an increase in the bias-conflict ratio within a dataset correlates with enhanced model performance. Therefore, employing data augmentation and facilitating the generation of bias-conflict samples without human intervention emerge as viable strategies for mitigating bias in models.

\begin{table}[h]
\centering
\caption{Accuracy of ResNet-18 classifier on benchmark test dataset. We report average of three trials on two bias-conflict ratios.}
\label{tab:vanilla_unbiased}
\resizebox{0.78\linewidth}{!}{

\begin{tabular}{lccc}
\hline
Dataset        & Bias-align ratio (\%) & Bias-conflict ratio (\%) & Accuracy (\%) \\ \hline
\multirow{2}{*}{CCIFAR-10} & 99.0   & 1.0   &    25.82           \\ 
                           & 95.0   & 5.0   &    39.42           \\ \hline
\multirow{2}{*}{BFFHQ}     & 99.0   & 1.0   &    60.96           \\ 
                           & 95.0   & 5.0   &    82.88           \\ \hline
\multirow{2}{*}{BAR}       & 99.0   & 1.0   &    29.70           \\ 
                           & 95.0   & 5.0   &    41.47           \\ \hline
\multirow{2}{*}{Dogs \& Cats} & 99.0 & 1.0   &   20.91            \\ 
                           & 95.0   & 5.0   &    33.20          \\ \hline
\end{tabular}
}
\end{table}

Previous works have attempted to tackle this problem through by either supervised learning or unsupervised learning. In a supervised learning setting, previous works define a re-weighting function~\cite{nam2020learning_LfF} or augment the features~\cite{kim2021biaswap, bahng2020learning_rebias} in the latent space, given that we know the true label of the input. In an unsupervised setting, we limit human guidance and debias the dataset. Previous unsupervised learning works~\cite{an20222,ko2023amplibias} use generative adversarial networks (GANs)~\cite{goodfellow2020generative} to generate bias-conflict samples and amplify the dataset. However, using GANs consumes the extensive amount of resources and takes a very long training time when trained from scratch. Furthermore, image translation using GANs requires careful tuning. Hence, we utilize pretrained diffusion models to save computational cost and time as we do not train the model.

Diffusion models~\cite{ho2020denoising,rombach2022high} have gained much attention due to its exceptional performance in image generations. Especially, pretrained diffusion models have generated realistic images based on given prompt. Hence, we propose a simple yet effective diffusion-based debiasing framework, coined~\sysname~for generating bias-conflict samples to mitigate dataset bias. To the best of our knowledge, we believe that this work is the first to utilize diffusion models in a debiasing task. Our framework leverages a pretrained generative model, a latent diffusion model, and an image-captioning model without any training these models. This design \textit{does not require any additional training} of generation models, rendering our approach both simplistic yet effective in synthesizing bias-conflict samples. Consequently, it facilitates debiasing the dataset and, in turn, the classifier.

In summary, the contributions of this paper are: (i). To the best of our knowledge, we are the first to leverage a pretrained latent diffusion model and image captioning model to generate synthetic bias-conflict samples for debiasing the classifier. (ii). Our work does not require training generative models, thus eradicating the learning cost. We utilize pretrained generative models, which are simple yet effective in synthesizing bias-conflict samples. (iii). Despite of no training cost, our model effectively amplified bias samples to tackle biased dataset, thus debiasing the classifier. 

%% file: main/03_related_work.tex
\section{Related Work}

Efforts to address dataset biases and enhance classifier fairness have proliferated in recent research, with methodologies broadly categorized into supervised and unsupervised debiasing techniques. Supervised debiasing typically focuses on mitigation without generating synthetic bias-conflict examples. Conversely, unsupervised approaches leverage generative modles to produce samples that explicitly confront and reduce biases.

\textbf{Debiasing via Supervised Methods} Early works employed supervised training with explicit bias labels~\cite{tartaglione2021end, sagawa2019distributionally, kim2019learning}. With predefined bias labels, previous works focused on extracting the bias features and attributes in the datgasests. Recent studies aim to address biases that do not rely explicitly on acquired bias labels. For instance, GroupDRO~\cite{sagawa2019distributionally} effectively clusters dataset subgroups under supervision, with subsequent enhancements introducing domain distribution generalizations through subgroup clustering and style mixing~\cite{goel2020model, zhou2021domain}. Methods such as LfF~\cite{nam2020learning_LfF} and Rebias~\cite{bahng2020learning_rebias} have further refined bias mitigation by employing dual models for bias conflict identification and feature disentanglement in latent spaces, respectively.

\textbf{Debiasing via Unsupervised Methods} explores generating or augmenting data to address the scarcity of bias-challenging examples. This includes innovative strategies like SelecMix~\cite{hwang2022selecmix}, which synthesizes such samples via mixup techniques, and BiasAdv~\cite{lim2023biasadv}, which employs adversarial attacks to alter bias cues in images. Generative Adversarial Networks (GANs) have also played a pivotal role, with various implementations aimed at retraining classifiers on amplified bias-conflict samples to reduce bias. Notably, A\textsuperscript{2}\cite{an20222} integrates StyleGAN2 for biased distribution learning, alongside few-shot adaptation techniques for sample augmentation\cite{ojha2021few}. AmpliBias~\cite{ko2023amplibias} further demonstrates the efficacy of FastGAN in a few-shot learning context for bias amplification.

Our work is inspired by these advancements and introduces a unique framework that capitalizes on a pretrained generative model, coupled with a latent diffusion model and an image-captioning model, to generate bias-conflict samples without additional training. This approach eliminates learning costs, offering a straightforward yet potent solution for reducing classifier bias.

%% file: main/04_methodology.tex
\section{Method}
We first describe the bias extraction methodology, where we extract bias-conflict samples from the training set without any human supervision. Then, we introduce our diffusion-based pipeline for generating bias-conflict samples based on the aforementioned extraction method. Finally, we discuss debiasing the classifier with the generated and original biased dataset.

\subsection{Problem Setup}

Within the domain of image classification, we explore a task that involves deducing the label of an input image, which is characterized by both intrinsic and bias attributes (e.g., intrinsic - animal, bias - color). Take, for instance, a biased training dataset \( \mathcal{I} = (x,y) \) encompassing attributes \( a_t, a_b \in \mathcal{A} \) representing intrinsic and bias attributes, respectively. The bias attribute \( a_b \) in the dataset drives the learning process to form correlations with the true label \( y \), thereby obstructing the model's aptitude in accurately discerning the intrinsic attribute \( a_t \). We elucidate two distinct scenarios within a biased dataset’s attribute \( a_b \): \textit{bias-aligned} samples, which are accurately classified under the unintended decision rule of the biased classifier \( f_B \)~\cite{nam2020learning_LfF}, and \textit{bias-conflict} samples, which are misclassified by the same rule of \( f_B \). Thus, the goal of this work is to transition the biased classifier \( f_B \) to a debiased classifier \( f_D \).

\begin{figure*}[!ht]
  \centering
  \includegraphics[width=0.8\linewidth]{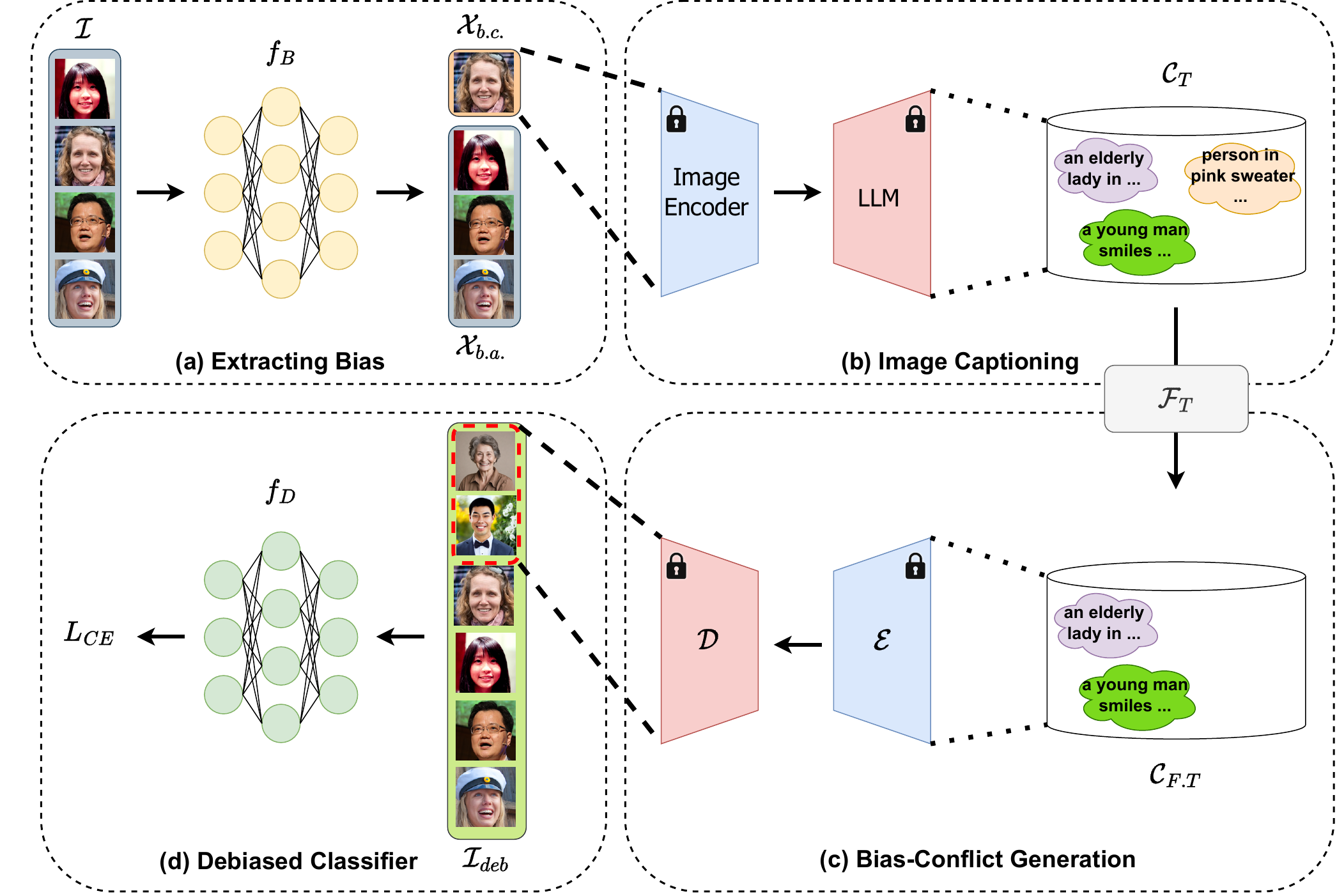}
  
  \caption{Our overall pipeline of~\sysname. It has four main components: (a), (b), (c), and (d), each executed sequentially. a) First, we intentionally train the classifier on the biased dataset and eventually make a biased classifier and extract bias based on top-$K$ losses. b) We utilize the extracted biased images and make image captions using freezed image encoder and LLM to create text corpus, $\mathcal{C}_T$. c) By leveraging pretrained latent diffusion model, we do text-to-image generation task to amplify bias by generating bias-conflict samples. d) Lastly, we use amplified dataset to train and debias the classifier.
}
    \label{fig:framework}
\end{figure*}

\subsection{Training a Biased Classifier}
In order to leverage the bias-conflict samples without necessitating human intervention, we deliberately train an image classifier to adopt a bias using the dataset \( \mathcal{I} \). As the model progressively learns from the bias-aligned samples, which are more prevalent in the dataset, it starts to exhibit a bias toward the attributes \( a_b \). To accentuate this biased prediction, we employ the Generalized Cross Entropy (GCE) loss~\cite{zhang2018generalized} in training the classifier, thereby favoring the bias-aligned samples, which are easier to learn. The GCE loss is formulated as follows:

\begin{equation}
\label{gce_loss}
    \mathcal{L}_{GCE} (p(x;\theta),y) = \frac{1-p_{y}(x;\theta)^q}{q},
\end{equation}

Given \( p(x;\theta) \) as the softmax output of the classifier and \( p_y(x;\theta) \) as the probability of the target attribute of the class \( y \), we introduce a hyperparameter \( q \in (0,1] \) to modulate the degree of amplification. Particularly, a value of \( q = 1 \) aligns the equation with the standard Cross-Entropy (CE) loss. The symbol \( \theta \) represents the learnable parameters of the classifier. Employing the GCE loss, we steer the model to adjust the weights of the gradients in favor of samples with higher probability \( p_y \), as written in the equation below:

\begin{equation}
    \frac{\partial {L}_{GCE} (p(x;\theta),y)}{\partial\theta} = p_y(x;\theta)^q \cdot \frac{\partial {L}_{CE} (p(x;\theta),y)}{\partial\theta},
\end{equation}

This inclines the model to exhibit higher losses for bias-conflict samples.

\subsection{Extracting Bias from Biased Classifier}
Upon training the biased classifier, we assume that samples exhibiting higher GCE losses are likely bias-conflict samples, given that the model has learned the bias attribute present in the biased dataset. Consequently, we extract the top-$K$ samples with elevated GCE losses from the training set, with $K$ set to 100 in this context. Despite certain datasets having a small proportion of bias-conflict samples in the training set (e.g., a bias-conflict ratio of 0.5\% or 1.0\%), we conjecture that $K$ value of 100 holds sufficient generalization potential to generate new bias-conflict samples. To elucidate further, consider an image of a cat from the Dogs and Cats dataset~\cite{kim2019learning}. The bias-aligned samples predominantly feature cats with ``black" or ``dark" fur, whereas the bias-conflict samples encompass cats with ``lighter or white fur''. Our aim, thus, is to extract images of light-furred cats from the biased classifier. The process of extraction of $\mathcal{X}_{b.c.}$ is formulated as follows:

\begin{equation} 
\begin{split}
& \mathcal{X}_{i...N} = \{x_\text{b.c.}, x_\text{b.a.}\}, \\
\text{where,} & ~~ x_{\text{b.c.}} = \argmax_{x_i \in D_{\text{biased}}} CE(f(x_i), y_i), \\ 
     \mathcal{X}_{{b.c.}} &= \{x_{{b.c.},1}, x_{{b.c.},2}, \dots, x_{{b.c.},K}\},
\end{split}
\end{equation}

where $x_{b.c.}$, $x_{b.a.}$ and $\mathcal{X}_{b.c.}$ symbolizes bias-conflict samples that have high CE losses, bias-align samples and a set of bias-conflict samples, respectively. Illustration of extracting bias from biased classifier can be found in Figure~\ref{fig:framework} (a).

\subsection{Image Captioning and Text-Filter}

\subsubsection{Image Captioning} Driven by the goal of minimizing human intervention in debiasing the classifier, we employ an image-captioning model to generate text captions based on the extracted bias-conflict samples, which will subsequently serve as the input source for text-to-image generation. In our setup, we adopt llava-llama-3-8b~\cite{2023xtuner}, a LLaVA~\cite{2024llava} model fine-tuned from LLaMA~\cite{touvron2023llama}, as our ``captioner'', leveraging a pre-trained image encoder and large language model (LLM). We utilize this LLM as a decoder-based language model to generate and predict ensuing text tokens, conditioned on a trained CLIP-ViT~\cite{radford2021learning}. Given a bias-conflict sample \(x_{b.c.}\) $\in \mathcal{X}_{b.c.}$, the captioner generates synthetic captions \(T_s\). To encapsulate a broader semantic understanding and features of the image through text, we generate \(three\) synthetic text captions, \(\{T_{s,1}, T_{s,2}, T_{s,3}\} \in \mathcal{C}_{T}\), which are collated in the text corpus \(\mathcal{C}_{T}\). We postulate that through this approach, the model can adeptly capture the semantics, class labels, and the bias-conflict attribute in the predicted text.
    
\subsubsection{Text-Filter} We employ a text filter,\( \mathcal{F}_{T} \), to enhance the quality of the generated text corpus, \( \mathcal{C}_{T} \), by filtering out noisy text through selection process of stop-words and top-$F$ words. Text-filter is illustrated in Fig.~\ref{fig:framework} from (b) to (c). Instead of allowing arbitrary generation, we strive to ensure that the text corpus aligns semantically with the classes in the dataset, denoted as \( \{C_{1}, C_{2}, ..., C_{n}\} \in \mathcal{C}_{N=1..n}\), where \( N \) and \( n \) represents the number of classes and the index of the last class, respectively. For instance, in the context of corrupted CIFAR10, all words are counted in the generated phrases/sentences. Then, we use stop words filter using NLTK library. This ensures to keep words of the class labels such as ``airplane'', ``automobile'', and ``boat'' which will appear more often then other objects. We choose top-$F$ frequent words where we count $2\times $ no. of classes. For example, for CCIFAR-10, $F=20$ since there are $10$ classes. Hence, we look at top-$20$ frequent words for our text filter $F_T$. Then, $F_T$ will filter out sentences that does not contain the top-$F$ words. This simple method would effectively filter out generated sentences that do not have relevance when augmenting bias-conflict samples. The final filtered text corpus is \( \mathcal{C}_{F.T} \)  $= \mathcal{F}_{T}(\mathcal{C}_{T})$. For example, in Fig.~\ref{fig:framework}, the caption ``person in pink sweater...'' is filtered since it does not contain class label of ``young'' or ``old''.

\subsection{Bias-Conflict Generation via Latent Diffusion Models}
Our primary generative model leverages Latent Diffusion Models (LDMs) based on Denoising Diffusion Probabilistic Models operating in the autoencoder's latent space. Initially, an autoencoder is trained on a substantial image dataset, where the encoder \( \mathcal{E} \) maps image \( x \) to a latent code \( z = \mathcal{E}(x) \). The decoder \( \mathcal{D} \) then reconstructs images from the latent code, approximating the original image \( x \) as \( \mathcal{D}(\mathcal{E}(x))\approx x \). Subsequently, an U-Net based diffusion model is trained to denoise the latent code, coupled with noise, and conditioned on various factors such as class labels, or text-embeddings. The LDM loss is forumlated by:


\begin{equation}
    L_{LDM} := \mathbb{E}_{\mathcal{E} (x),y,\epsilon\sim\mathcal{N}(0,1),t}\left [ \left \| \epsilon-\epsilon_{\theta}(z_t,t,c_{\theta}(y)) \right \|^2_2 \right ]
\end{equation}

where \( c_\theta(y) \) is the condition vector, predicated on \( y \). \( z_t \) represents the latent code with noise at time \( t \), while \( \epsilon_\theta \) and \( \epsilon \) denote the denoising network and the noised sample of \( x \), respectively. In our work, we condition text prompts for the condition vector \( c_\theta(y) \) to generate denoised image $\tilde{x}$.

Our methodology is summarized in Fig.~\ref{fig:framework}. In essence, we deliberately train a biased classifier, \( f_B \), on a biased dataset. Following this, we extract bias-conflict samples, \( x_{b.c.} \), from the trained biased classifier employing the CE loss, facilitating the identification of samples with high CE loss, synonymous with bias-conflict samples. Thereafter, a captioner is deployed to generate synthetic captions for the bias-conflict samples, which are aggregated in the text corpus \( \mathcal{C}_T \) and subsequently refined through a text filter, \( \mathcal{F_T} \), to generate filtered text corpus $\mathcal{C_{F.T}}$. Next, we generate denoised image $\tilde{x}$, conditioned on the text prompts from the corpus $\mathcal{C_{F.T}}$ using latent diffusion model. Lastly, we create a debiased dataset $\mathcal{I}_{deb}$ where $\mathcal{I}_{deb} = \mathcal{X}_{b.c.} + \mathcal{X}_{b.a.} + \mathcal{X}_{generated}$, where we use it to train the biased classifier to eventually become a debiased classifier, $f_D$.

%% file: main/05_experiments.tex
\section{Experiments}
\subsection{Experimental Setup}
\subsubsection{Datasets}~\label{dataset_appendix}The intrinsic and bias attributes of these datasets are enclosed in brackets as \{intrinsic, bias\}. Specifically, the attributes for the datasets used in our experiments are delineated as follows: CCIFAR10~\cite{hendrycks2019ccifar10}: \{object, noise\}, BFFHQ~\cite{kim2021biaswap}: \{age, gender\}, Dogs \& Cats~\cite{kim2019learning}: \{animal, color\}, and BAR~\cite{nam2020learning_LfF}: \{action, background/environment\}. Each dataset encompasses a spectrum of bias-conflict sample ratios, ranging from 0.5\% to 5\%. The available bias-conflict sample ratios are 0.5\%, 1.0\%, 2.0\%, and 5.0\% for CCIFAR10 and BFFHQ, while Dogs \& Cats and BAR have 1.0\% and 5.0\% available.

\subsubsection{Implementation Details} In accordance with the protocol outlined in~\cite{nam2020learning_LfF}, we employ a ResNet-18 architecture~\cite{he2016deep} across all datasets. Notably, the BAR dataset, owing to its relatively small number of images compared to other benchmark datasets, necessitates the use of a pretrained ResNet-18 when training the classifier. Conversely, for the remaining datasets, we trained the models from scratch. Each classifier undergoes a training for 50 epochs. We report the average over three independent trials for test set accuracy. For GCE loss, we set $q = 0.7$ equivalent to the experimental setting of LfF~\cite{nam2020learning_LfF} for fair comparison.

For fair comparison of measuring carbon emission and time taken for efficiency of generative model, we used the same hardware settings for all experiments. We used 1 $\times$ A100 GPU, 16 core AMD EPYC 7763 64-Core Processor CPU and 16GB RAM for all experiments and ablation study.

\begin{table}[!t]
\caption{Example text prompts captioned from extracted bias-conflict samples. For each dataset, we illustrate three example texts. Important words that represent the classes of the dataset is highlighted in bold.}
\label{text_captions_generated}
\centering
\resizebox{1.0\linewidth}{!}{

\begin{tabular}{|c|cll|}
\hline
\textbf{Datasets} & \multicolumn{3}{c|}{\textbf{Examples}}                                                                                                                                                                                                                                       \\ \hline
CCIFAR10          & \multicolumn{3}{c|}{\begin{tabular}[c]{@{}c@{}}an \textbf{airplane} sits on the tarmac with several cars surrounding it\\ a \textbf{frog} with its eyes closed sits on top of a brown rock\\ two white \textbf{horses} and a red horse standing next to each other in an empty room\end{tabular}}          \\ \hline
BFFHQ             & \multicolumn{3}{c|}{\begin{tabular}[c]{@{}c@{}}a \textbf{young woman} in a blue shirt and black shorts with her smiling face\\ an \textbf{asian woman} in a yellow blouse standing next to some jewelry\\ the \textbf{man} wearing the hat with the bird on it sits in front of green leaves\end{tabular}} \\ \hline
BAR               & \multicolumn{3}{c|}{\begin{tabular}[c]{@{}c@{}}the man is climbing up the side of an \textbf{ice wall}\\ a lady in an orange shirt is holding a \textbf{fishing pole} while standing on a boat\\ a male athlete jumps over a pole in a \textbf{pole vault} competition\end{tabular}}                       \\ \hline
Dogs \& Cats             & \multicolumn{3}{c|}{\begin{tabular}[c]{@{}c@{}} \textbf{a black dog} sitting on top of a person's car\\ \textbf{a yellow cat} is in a cage inside an animal shelter\\ a man is holding up a small \textbf{brown puppy}\end{tabular}} \\ \hline

\end{tabular}
}
\end{table}

\begin{table*}[!t]
\setlength{\tabcolsep}{4pt}
  \caption{Performance comparison on image classification task on unbiased test datasets across four benchmark datasets. The best and runner-up performance is in bold and underlined, respectively.}
  \label{tab:performance}
  \centering
  \resizebox{1.0\linewidth}{!}{
  
  \begin{tabular}{ccccc|cccc|cc|cc}
    \toprule
    \multirow{3}{*}{Methods} & \multicolumn{4}{c}{Synthetic} & \multicolumn{8}{c}{Real-world} \\
    \cmidrule(r){2-5} \cmidrule(r){6-13}
    & \multicolumn{4}{c|}{Corrupted CIFAR-10}     &  \multicolumn{4}{c|}{BFFHQ} & \multicolumn{2}{c|}{Dogs \& Cats } & \multicolumn{2}{c}{BAR} \\
    & 0.5\% & 1.0\% & 2.0\% & 5.0\% &0.5\% & 1.0\% & 2.0\% & 5.0\% & 1.0\% & 5.0\%& 1.0\% & 5.0\% \\
    \midrule
    Vanilla~\cite{he2016deep} & 23.08& 25.82& 30.06 & 39.42 & 55.64&60.96&69.00&82.88&48.06&69.88&70.55&82.53\\
    LfF~\cite{nam2020learning_LfF} & \underline{29.40}& 33.57&38.39&49.47&66.50&69.89&74.27&78.31& 71.72& 84.32& 69.92& 82.92\\
    Rebias~\cite{bahng2020learning_rebias} & 21.45&23.62&30.12&41.68&55.76&60.68&70.13&83.36& 48.70&65.74 &73.01&83.51\\
    DisEnt~\cite{lee2021learning_disent} & 28.61& \underline{36.49}&\underline{42.33}&51.88&62.08&66.00&69.82&80.68& 65.74&81.58 &69.49&82.92 \\
    LfF+BE$^*$~\cite{lee2022biasensemble} & 27.94& 34.77& 42.12& \underline{52.16}& 67.36& 75.08& 80.32& 85.48& \underline{81.52}& \underline{88.60}& \underline{73.36}& 83.87\\
    $A^2$~\cite{an20222} & 23.37&27.54&30.60&37.60&77.83&78.98&81.13&86.22& - & - &71.15&83.07\\
    AmpliBias~\cite{ko2023amplibias} &
     \textbf{34.63} & \textbf{45.95} & \textbf{48.74} & \textbf{52.22} & \underline{78.82} & \textbf{81.80} & \underline{82.20} & \underline{87.34} & 73.51 & 83.95 & 73.30& \underline{84.67}\\
    \bottomrule
    \sysname & 27.73 &30.82&31.83&40.93&\textbf{78.88}&\underline{79.48}& \textbf{83.20}& \textbf{88.31} & \textbf{84.17}& \textbf{94.33}& \textbf{74.35}& \textbf{85.23}\\
    \bottomrule
    \multicolumn{13}{c}{\small $\ast$ denotes that the results of the method are obtained from the official paper of the method.} \\
  \end{tabular}
  }
\end{table*}

\begin{figure*}[h]
  \centering
  \includegraphics[width=1\linewidth]{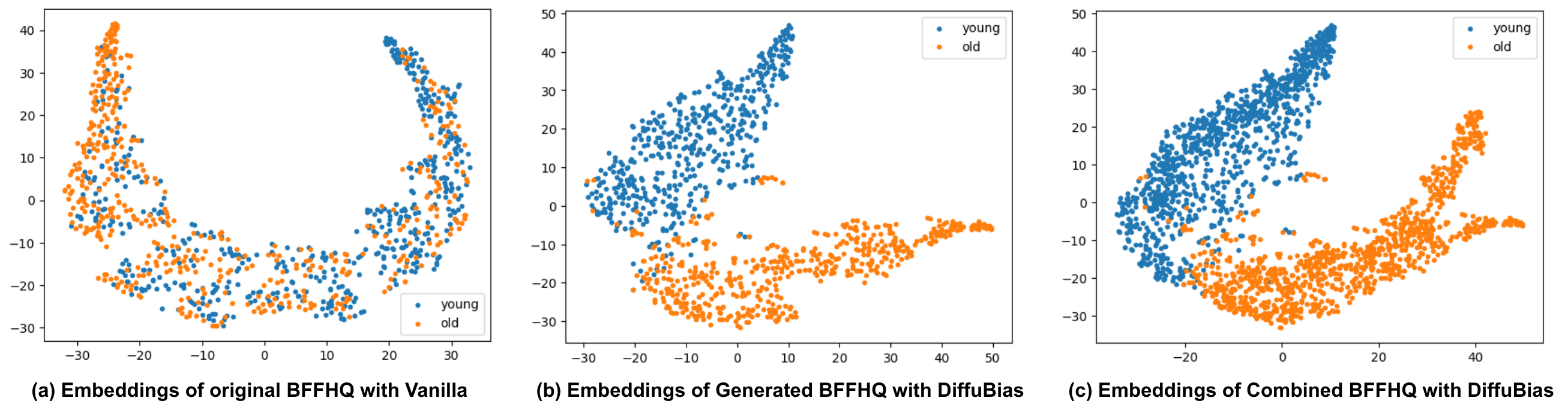}
  
  \caption{t-SNE embeddings of BFFHQ using vanilla ResNet-18 and DiffuBias of label young and old. (a) demonstrates that original BFFHQ dataset is heavily biased thus making the vanilla ResNet-18 difficult to classify given the scarce number of bias-conflict samples. (b) and (c) demonstrates t-SNE embeddings of generated and generated + original dataset, respectively.}
  \label{tsne}
\end{figure*}

\subsection{Performance Evaluation}
\subsubsection{Performance Evaluation in Contexts with High Bias-Conflict Ratios}

The analysis of our experimental results, as detailed in Table~\ref{tab:performance}, demonstrates that our model exhibits superior performance in scenarios characterized by high bias-conflict ratios. 
The generation process involves creating text prompts that encapsulate bias-conflict attributes, thereby enhancing the production of bias-conflict samples. 
Illustrative examples of text prompts and the corresponding generated images are presented in Table~\ref{text_captions_generated}.

Notably, our model achieved its highest performance on the Dogs \ Cats dataset at a 5.0\% bias-conflict ratio, recording an accuracy of 94.33\%. This represents a significant improvement of 5.73\% over the combined LfF + BE approach. In the case of the BFFHQ dataset, our model also outperformed baseline models, achieving a leading accuracy of 88.31\% with a 0.97\% difference from the runner-up performance. The observed performance advantage is primarily attributed to the model's ability to accurately interpret and utilize the semantics of bias-conflict features. This capability proves particularly advantageous in comparison to synthetic datasets like CCIFAR-10, where noise elements present more considerable challenges to distinction, even to human observers. Our results suggest that the nuanced understanding and generation of text captions that accurately reflect bias-conflict attributes play a critical role in enhancing model performance in complex bias scenarios.


\subsubsection{Analysis of underperformance in synthetic dataset} Our experiments demonstrated a suboptimal performance of our model on the CCIFAR-10 dataset. We believe that this shortfall is due to the high resolution in the generated samples. Generated high-resolution images enable enhanced semantic comprehension and feature extraction during training. However, the model faces challenges with small resolution of test images. Since we align our experimental setup with prevailing benchmarks by downscaling images to \(32 \times 32\), a significant loss of semantic information and distinctive features was observed, diminishing the model's efficacy. Additionally, the synthetic bias-conflict inherent in the CCIFAR-10 dataset's test set posed a challenge. We intend the image captioning model to process texts and include classes such as "deer" or "horse" and bias-conflict attributes such as "noise" or "blur". However, it was not able to capture the intended bias-conflict attributes occasionally during caption generation that led to an inadequately trained model on bias-conflict attributes, further impacting its performance during testing.

\subsubsection{Synthetic bias-conflict samples reflecting real-world samples} Our empirical analysis underscores the robustness of our framework when applied to real-world datasets, attaining state-of-the-art performance across diverse ratios within three benchmark datasets. For instance, at a \(5\%\) ratio, our model scored accuracies of \(88.31\%\), \(94.33\%\), and \(85.23\%\) on BFFHQ, Dogs \& Cats, and BAR datasets, respectively. This suggests that the generated images bear a close resemblance to real-world scenarios. 
A notable observation is the significant outperformance on the Dogs and Cats dataset, compared to other datasets, which were improvements of about \(2\%\). We postulate that LDMs have advantages at synthesizing animal and facial images compared to other domains images with dynamic movements or unusual images. 

\begin{figure}{r}
  \centering 
  \includegraphics[width=0.48\textwidth]{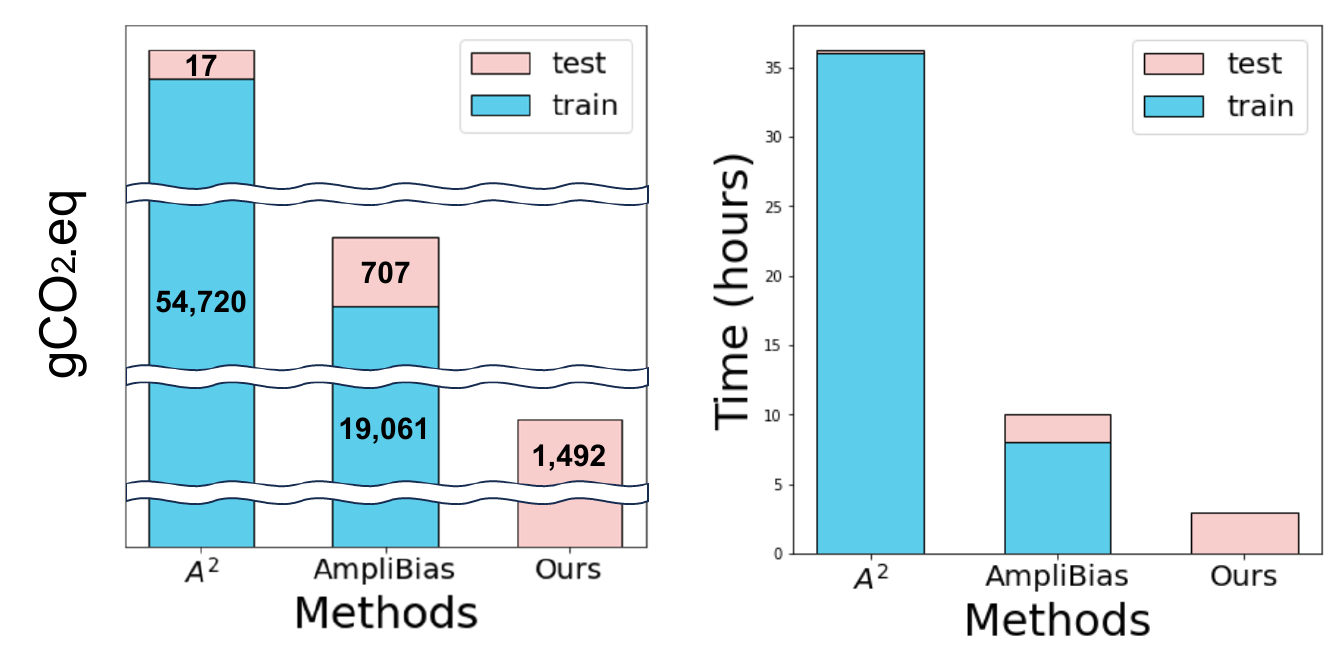}
  \caption{Comparative analysis of generative models based on carbon emissions gCO\textsubscript{2}.eq and computational time (hours) for the BFFHQ dataset. The graphs illustrate gCO\textsubscript{2}.eq emissions and hours respectively, with disproportionate scales for clarity.}
  \label{fig:carbon_time}
  \vspace{-10pt} 
\end{figure}

\begin{figure*}[h]
  \centering
  \includegraphics[width=1\linewidth]{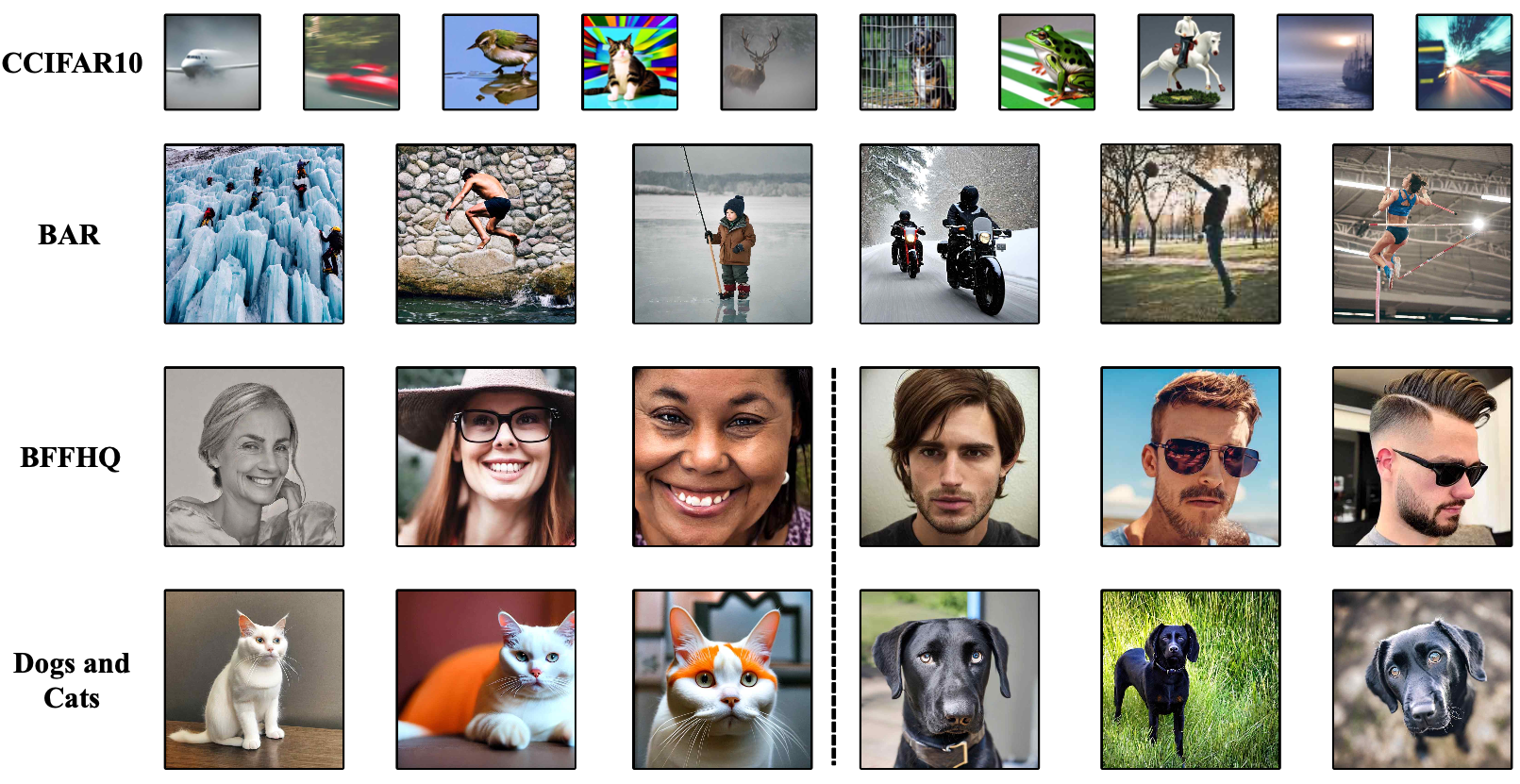}
  \caption{Generated synthetic bias-conflict samples from our proposed framework, DiffuBias.}  
  \label{biasconflict}
\end{figure*}

  

\begin{figure*}[htbp!]
  \centering
  \includegraphics[width=0.8\linewidth]{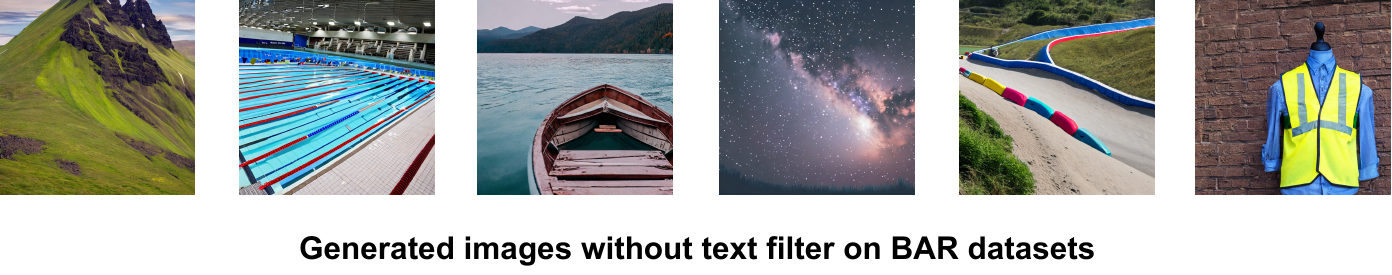}
  \caption{Wrong images generated without text filter, $\mathcal{F_T}$ from BAR captions. The images do not contain any actions. Instead, only background (task-irrelevant feature) part of the text is generated.}
  \label{without_filter_imgs}
\end{figure*}

\subsubsection{Carbon Emission and Energy Consumption Analysis.}
We analyze carbon emission and energy consumption for two GAN-based models ($A^2$ and AmpliBias) compared with a diffusion-based generative model for train and test in BFFHQ dataset. Carbon dioxide emissions are calculated through the CodeCarbon library~\cite{codecarbon}. CO\textsubscript{2}eq emissions are determined by the formula Carbon Intensity (C) × Energy consumed during computation (E), using the average global carbon intensity value of 475 gCO\textsubscript{2}eq/KWh to ensure a fair comparison. As illustrated in Fig.~\ref{fig:carbon_time}, our model required a total of 3 hours to generate an equivalent number of bias-conflict samples to the bias-aligned samples. In contrast, $A^2$ and AmpliBias required over 35 and 10 hours, respectively, for training and inference. During the inference phase, our model consumed more electricity, resulting in higher CO\textsubscript{2} emissions. This increased consumption is attributable to the simultaneous computation of text generation (image captioning) and text-to-image generation steps. Moreover, diffusion models generally have more parameters than GANs, thereby incurring higher computational costs during the inference step. Nevertheless, overall, our model demonstrates lower energy consumption and CO\textsubscript{2} emissions. The results indicate that pretrained diffusion models not only possess robust feature space representations but also conserve energy and reduce CO\textsubscript{2} emissions, thus effectively debiasing the biased classifier.

\subsection{Discussion and Further Analysis}

\subsubsection{t-SNE Analysis} As illustrated in Fig.~\ref{tsne}, the t-SNE embeddings from the pretrained vanilla ResNet-18 model fail to distinctly cluster the label attributes, young and old. Conversely, our trained DiffuBias framework successfully clusters these two attributes, indicating that the generated bias-conflict samples aid in debiasing the classifier. This analysis is further supported in sub-figure (c), where the combination of generated bias-conflict samples and original BFFHQ datasets maintains a clear separation within the t-SNE embedding space. Thus, these results demonstrate that training with the generated dataset effectively debiases the classifier, underscoring the practical utility of our approach.

\begin{figure*}[h]
  \centering
  \includegraphics[width=0.60\linewidth]{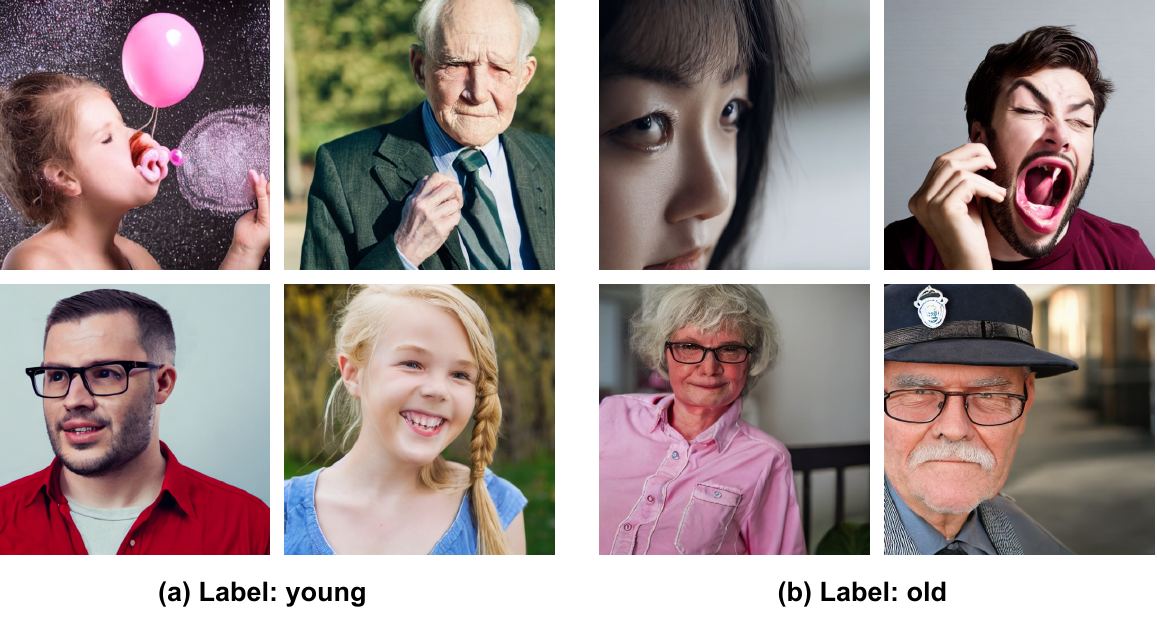}
  
  \caption{Examples of generated images using DiffuBias pipeline of BFFHQ dataset. The first row of images are samples generated that mismatches with the given label. The second row of images are correctly generated samples, following the label.}
  \label{generated_with_filter}
\end{figure*}

\subsubsection{Generated Bias-Conflict Images}
In Figure~\ref{generated_with_filter}, we present a selection of bias-conflict samples generated by our model. Certain generated images either do not correspond with the provided labels or exhibit incompletely rendered features. For instance, under the label "young," the depiction of a young girl with a balloon showcases incomplete rendering of the mouth and nose. Despite this, it is still discernible that the image represents a young girl. Additionally, a misrepresentation is observed in the generation of an old man dressed in a suit. This anomaly could stem from the extraction process where images of an old man were erroneously categorized as bias-conflict samples. In a similar vein, under the label "old," the images in the first row do not align with the given label, possibly a result of erroneous extraction as bias-conflict samples during the dataset preparation stage.

Figure~\ref{without_filter_imgs} illustrates synthesized images without text filters in BAR dataset. When filters are not used, we generate images without human actions. Instead, the images do not contain any actions but the bias attributes (background). This result reinforces the importance of text filters when generating bias-conflict images via text prompts.

\begin{figure}[htbp]
  \centering
  \includegraphics[width=1.0\linewidth]{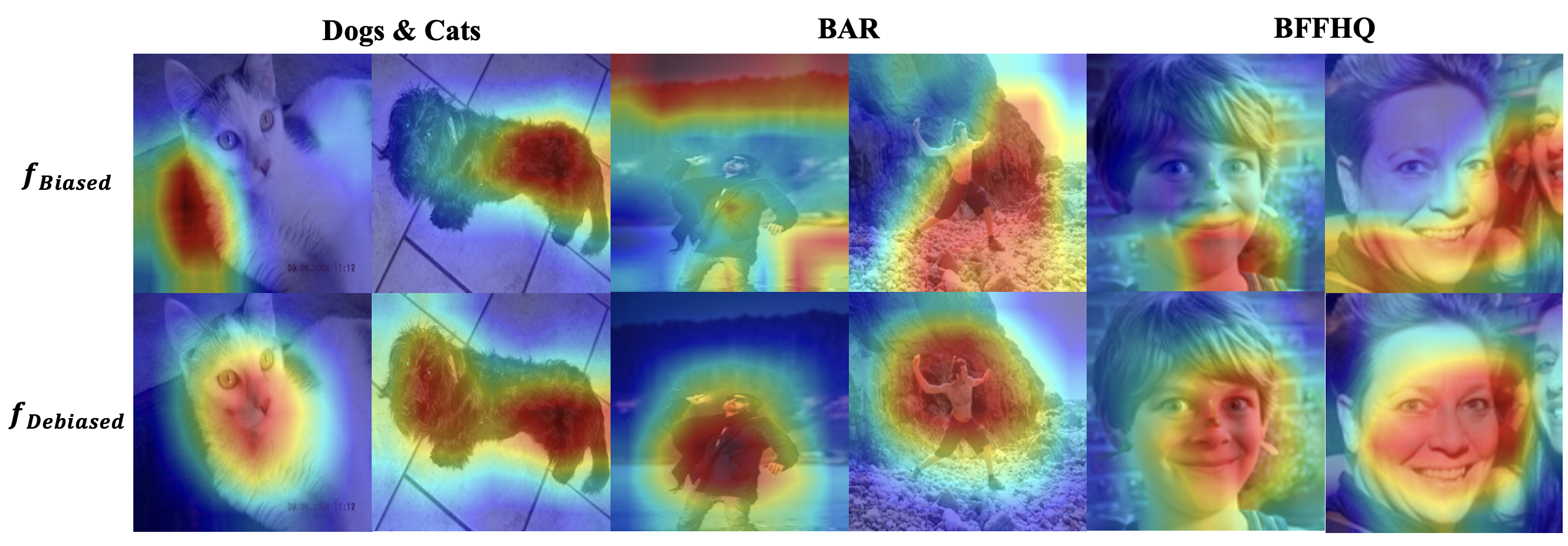}
  \caption{GradCAM results of $f_{Biased}$ and $f_{Debiased}$ for three benchmark test dataset. The test set is often bias-conflict samples which are difficult for the model to predict since the vanilla model is biased.}
  \label{gradcam_results}
\end{figure}

\subsubsection{Text Prompts Quality Analysis} In Table~\ref{text_captions_generated}, we present filtered text captions from the text corpus, utilized for generating synthetic bias-conflict samples. For each dataset, three text captions were selected from the top-$K$ extracted loss samples. The results in Table~\ref{tab:performance} demonstrate that our framework exhibits superior performance on real-world datasets, a finding that is substantiated by the example text prompts provided. It is evident that text captions from BFFHQ and BAR exhibit features indicative of bias-conflict, such as "young woman", "ice wall", and "pole vault". Conversely, for CCIFAR10, the model struggles to capture the bias-conflict setting, which in this instance pertains to the type of noise present. Given the small size of the images, it is challenging to ascertain the specific type of noise applied. Consequently, the pretrained captioner prioritizes class label identification, underperforming on synthetic datasets.

\subsubsection{GradCAM Visualization and Analysis}
We sought to visualize and understand the mechanisms through which the networks \(f_B\) and \(f_D\) interpret bias-conflict images, especially when predicting their true class labels is challenging due to the presence of task-irrelevant features in the samples through GradCAM~\cite{selvaraju2017grad} visualization. Figure~\ref{gradcam_results} presents visualizations across three real-world datasets. In the Dogs \& Cats dataset, \(f_B\) predominantly focuses on fur color, associating it with class labels—this is attributed to the dataset's bias towards ``black cats'' and ``white dogs''. Conversely, \(f_D\) concentrates on the animals themselves, i.e., the task-relevant features, thereby enabling accurate class prediction. In the BAR dataset, although \(f_B\) recognizes the importance of action, it erroneously also emphasizes the background—a task-irrelevant feature—thus diminishing the biased classifier's performance. In contrast, \(f_D\) remains attentive solely to the action performed by the subjects, avoiding distraction by irrelevant background features. 
GradCAM visualization for the BFFHQ dataset reveals intricate patterns. The network $f_D$ predominantly focuses on the central facial region when determining the class (young or old), which we surmise is due to the model's reliance on detecting wrinkles and skin texture—a common indicator of aging. Conversely, $f_B$ appears to concentrate on the chin and peripheral facial areas, where signs of aging such as wrinkles or a double chin are prominent. This too could be construed as a logical approach to age classification. Nonetheless, it is $f_D$ that demonstrates superior performance in classifying bias-conflict samples, likely due to its more holistic analysis of facial textures, including the skin.

\subsubsection{Further Analysis of Generated Samples}
We demonstrate examples of generated samples from the four benchmark datasets in Figure~\ref{biasconflict}. Our proposed DiffuBias were able to generate images based on the difficult bias-conflict conditions such as frost, haze and blur generated in the airplane, deer and automobile, respectively. Furthermore, in Dogs \& Cats dataset, our framework was able to generate ``bright fur cats'' and ``dark fur dogs'' which are expected bias-conflict pairs.
This demonstrates that our framework was able to capture the intended bias-conflict representation only with the pre-trained captioner. Thus, leveraging pre-trained captioner allows us to generate images without any training.

%% file: main/06_discussion.tex
\begin{table}[h]
\caption{Ablation study results on our method with and without text filter for real-world dataset. Best performance is in bold.}
\label{tab:ablation}
\centering
\resizebox{1.0\linewidth}{!}{

\begin{tabular}{lcccccccc}
\toprule
Methods & \multicolumn{4}{c}{BFFHQ} & \multicolumn{2}{c}{BAR} & \multicolumn{2}{c}{Dogs \& Cats} \\
\cmidrule(lr){2-5} \cmidrule(lr){6-7} \cmidrule(lr){8-9}
& 0.5\% & 1.0\% & 2.0\% & 5.0\% & 1.0\% & 5.0\% & 1.0\% & 5.0\% \\
\midrule
\sysname & 78.88 & 79.48 & \textbf{83.20} & \textbf{88.31} & \textbf{74.35} & \textbf{85.23} & \textbf{73.51} & \textbf{83.95} \\
\sysname~w/o text filter & \textbf{78.94} & \textbf{80.67} & 82.87 & 86.29 & 66.43 & 67.66 & 62.71 & 68.39\\
\bottomrule
\end{tabular}
}
\end{table}

\subsubsection{Impact of Text-Filter on Performance} 
In this section, we explore the effect of text-filtering by conducting an ablation study on the presence of text-filter. We compared the performance when the text filter is not applied and bias-conflict samples are generated directly. Table~\ref{tab:ablation} illustrates that~\sysname~without text filter underperforms in BFFHQ, BAR and Dogs \& Cats datasets. Particularly, for BFFHQ dataset, bias-conflict ratio levels of 2\% and 5\% have a marginal difference of $82.87\%$ and $86.29\%$ in comparsion with other datasets. In BFFHQ dataset, we observe a better performance at higher bias-conflict ratios as more bias-conflict samples are extracted. However, the impact is less pronounced at 0.5\% and 1.0\% ratios as DiffuBias without text filter has slightly better performance with accuracy of $78.94\%$ and $80.67\%$, respectively. This result indicates that the generated images have a lesser effect in debiasing the model in these cases. Furthermore, GradCAM results reinforce that there are many complex attributes that determines the age of a person's face. On the other hand, the significance of the text filter is more apparent in the BAR and Dogs \& Cats datasets where DiffuBias with text-filter outperforms in all percentage of bias-conflict samples. Given that the text prompts provide a detailed description of actions in BAR dataset and fur colors of animals in Dogs \& Cats dataset, filtering is necessary before proceeding to image synthesis. Thus, text filter aids in enhancing model performance across different datasets.

%% file: main/07_conclusion.tex

\subsection{Limitations \& Future Work} A limitation inherent to our work is the sole reliance on pretrained models, which may confine our approach to certain domains or tasks. For instance, our model adeptly generates images of animals or humans, yet may encounter difficulties with medical images if biases are present within them. Additionally, tasks with unbalanced labels or classes, such as anomaly detection, pose a challenge, particularly in generating anomaly images. 

Thus, in our future work, we aim to extend our work on fine-tuning pretrained diffusion models to become adaptive to several domains in order to mitigate bias in the dataset eventually enhancing performance. Also, this would help in better performance in low-resolution image scenarios, such as CCIFAR-10 dataset. 

\section{Conclusion}
In this work, we introduce~\sysname, a debiasing framework that leverages the amplification of bias to mitigate its effects. Our framework is founded on the key insight that amplifying bias-conflict samples facilitates the learning of task-relevant features, thereby reducing bias. 
Results demonstrate that our methods achieve state-of-the-art performance for most of the benchmark dataset. In addition, our model was able to achieve faster training and inference times compared to baselines that utilize generative models. We believe that our framework and findings provide valuable insights into the significance of augmenting and amplifying the bias of the dataset, thereby establishing a stronger foundation in addressing bias in the field of computer vision.

%% file: main/appendix.tex
\appendix
\section{Appendix}

\begin{figure*}
  \centering
  \includegraphics[width=1\textwidth]{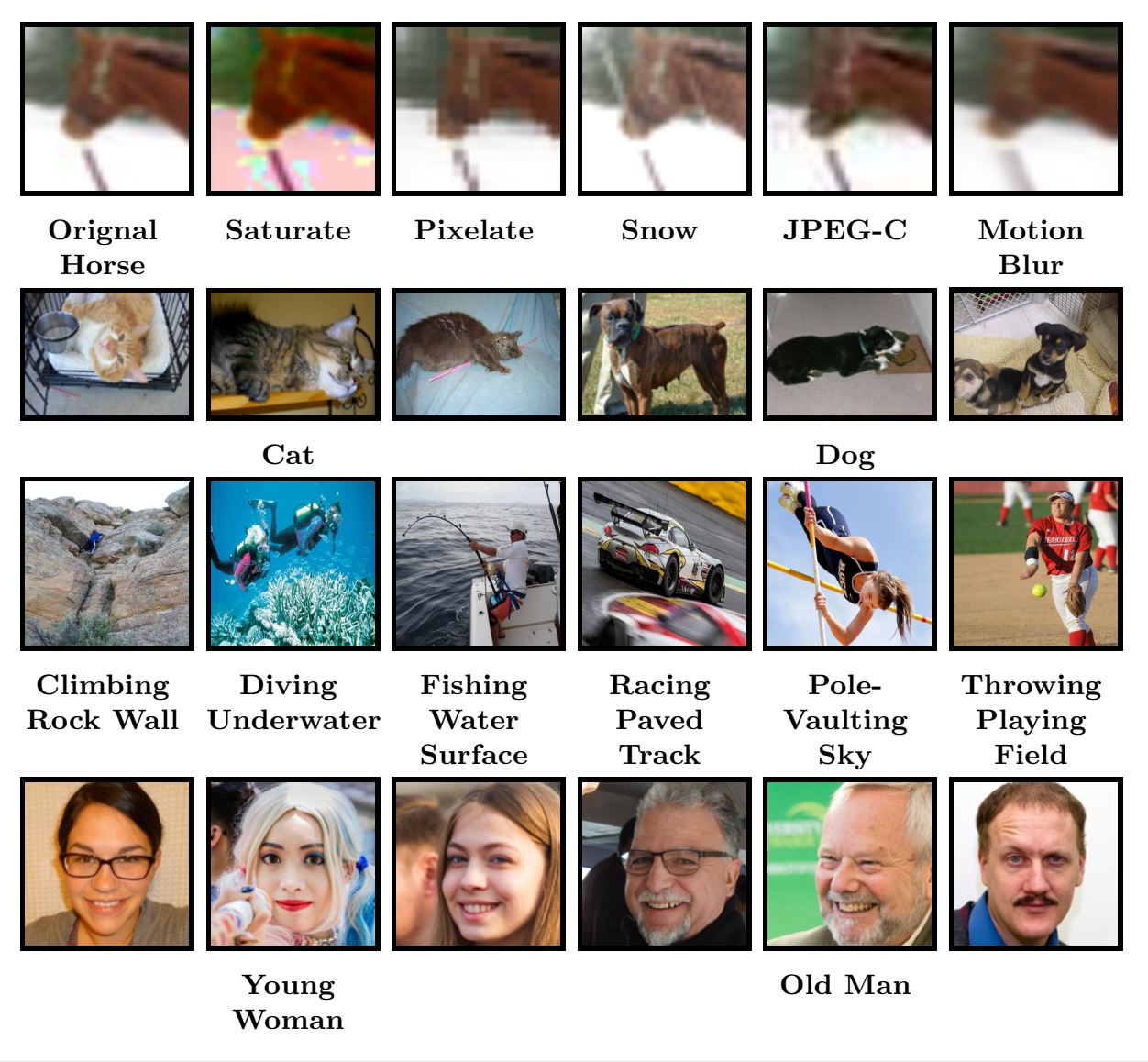}
  \caption{Example images of biases present in benchmark dataset. From the top, the datasets are CCIFAR10, Dogs \& Cats, BAR and BFFHQ, respectively.}
  \label{dataset_explanation}
\end{figure*}

\subsection{Datasets}
We elaborate further on the details for the benchmark datasets used in our experiments. Details of the dataset composition can be found in Table~\ref{dataset_composition}. Furthermore, we demonstrate examples of bias images in Fig.~\ref{dataset_explanation} for the four benchmark datasets. From the top, CCIFAR-10, Dogs \& Cats, BAR, and BFFHQ bias-align images with their intrinsic attributes can be found, respectively. \\
\textbf{Corrupted CIFAR-10 (CCIFAR10).}~\cite{hendrycks2019ccifar10} matched each class from original CIFAR-10 with a single noise corruptions as the dominant bias (bias-align) in the dataset. The unique corruption pairs of image class-noise are as follows: ``Airplane-Snow'', ``Automobile-Frost'', ``Bird-Defocus Blur'', ``Cat-Brightness'', ``Deer-Contrast'', ``Dog-Spatter'', ``Frog-Frosted Glass Blur'', ``Horse-JPEG Compression'', ``Ship-Pixelate'', ``Truck-Saturate''. Bias-conflict samples are composed of the class with noises that was not paired as bias-aligned samples. For example, ``Airplane'' has bias-conflict samples of other noises such as ``Frost, Spatter, Saturate''. 

\textbf{Biased Flickr-Faces-HQ (BFFHQ).}~\cite{lee2021learning_disent} is a benchmark dataset for debiasing based on FFHQ~\cite{karras2020stylegan2} dataset. BFFHQ dataset's classes are ages ''Old'' and ''Young''. Their correlating biases are ''Male'' and ''Female'' where bias-align samples are ''old men'' and ''young women'', respectively. Since age range is very ambiguous, old men are selected between 40 to 59 years old while young women images are gathered ranging from 10 to 29 years old. \\

\textbf{Biased Action Recognition (BAR).}~\cite{nam2020learning_LfF} is a dataset gathered from imSitu, Standford 40 Actions and Google Image Search with six different actions as classes with various backgrounds, reflecting the real-world setting. Six classes with their matching bias backgrounds are as follows: ``Climbing-Rock Wall'', ``Diving-Underwater'', ``Fishing-Water Surface'', ``Racing-A Paved Track'', ``Throwing-Playing field'' and ``Valuting-Sky''. 

\textbf{Dogs \& Cats}.~\cite{kim2019learning} is a dataset extracted from original Dogs and Cats dataset in Kaggle. The original dataset is composed of 25,000 train images and 12,500 test images. There are two classes (dogs and cats) of images matched with their fur color as follows: ``dog-bright colored fur'' and ``cat-dark colored fur''.

\newcolumntype{P}[1]{>{\centering\arraybackslash}p{#1}}

\begin{table*}[htbp]
\centering
\begin{tabular}{|c|c|c|c|c|c|}
\hline
\textbf{Dataset} & \textbf{Bias Ratio} & \multicolumn{2}{c|}{\textbf{Training}} & \multicolumn{2}{c|}{\textbf{Test}} \\ \cline{3-6} 
                 &                     & \textbf{Bias-align} & \textbf{Bias-conflict} & \textbf{Bias-align} & \textbf{Bias-conflict} \\ \hline
\textbf{CMNIST}  & 0.5\%               & 54,729              & 271                    & 1,006               & 8,994                  \\
                 & 1.0\%               & 54,454              & 546                    &                     &                        \\
                 & 2.0\%               & 53,904              & 1,096                  &                     &                        \\
                 & 5.0\%               & 52,254              & 2,746                  &                     &                        \\ \hline
\textbf{CCIFAR10}& 0.5\%               & 44,832              & 228                    & 1,000               & 9,000                  \\
                 & 1.0\%               & 44,527              & 442                    &                     &                        \\
                 & 2.0\%               & 44,145              & 887                    &                     &                        \\
                 & 5.0\%               & 42,820              & 2,242                  &                     &                        \\ \hline
\textbf{BFFHQ}   & 0.5\%               & 19,104              & 96                     & 500                 & 500                    \\
                 & 1.0\%               & 19,008              & 192                    &                     &                        \\
                 & 2.0\%               & 18,816              & 384                    &                     &                        \\
                 & 5.0\%               & 9,120               & 960                    &                     &                        \\ \hline
\textbf{BAR}     & 1.0\%               & 1,761               & 14                     & N/A                 & 570                    \\
                 & 5.0\%               & 1,761               & 85                     &                     &                        \\ \hline
\textbf{Dogs \& Cats} & 1.0\%          & 8,037               & 80                     & N/A                 & 1,000                  \\
                      & 5.0\%          & 8,037               & 452                    &                     &                        \\ \hline
\end{tabular}
\caption{Dataset composition for benchmark datasets. Bias ratio indicates the proportion of bias-conflict samples present in the total training data.}
\label{dataset_composition}
\end{table*}

\subsection{Training Configuration}
Table~\ref{model_configuration} demonstrates the training configuration used in our experiments. We used different configuration for synthetic dataset (CCIFAR10) and real-world dataset (BFFHQ, BAR and Dogs \& Cats). Furthermore, this different configuration comes from the different image size where CCIFAR10 has a very small image size compared to other three real-world datasets. BAR and Dogs \& Cats datasets leveraged pretrained ResNet18 on ImageNet datasets as the number of dataset is very small as it can be seen in Table~\ref{dataset_composition}. Moreover, we aligned our experimental settings with previous debiasing works~\cite{nam2020learning_LfF, bahng2020learning_rebias, lee2022biasensemble, lee2021learning_disent, kim2019learning, an20222, ko2023amplibias} for fair comparison.
\begin{table*}[ht]
\centering
\resizebox{\textwidth}{!}{
\begin{tabular}{|l|c|c|l|l|c|}
\hline
\multicolumn{1}{|c|}{\textbf{Dataset}}   & \textbf{Image Size} & \textbf{Batch Size} & \multicolumn{1}{c|}{\textbf{Scheduler}} & \multicolumn{1}{c|}{\textbf{Augmentation}}  & \textbf{ImageNet Pretrained} \\
                   &                     &                     & \multicolumn{1}{c|}{\textbf{(Decay Rate)}} &                                            &                              \\ \hline
\multicolumn{1}{|c|}{CCIFAR10}           & 32 x 32             & 256                 & \multicolumn{1}{c|}{0.5 decay}                                &      \multicolumn{1}{c|}{\multirow{2}{*}{X}}                 & \multirow{2}{*}{X}                            \\
                   &                     &                     & (every 20 epochs)                        &                                 &                              \\ \hline
\multicolumn{1}{|c|}{BFFHQ}             & 224 x 224           & 64                  & \multicolumn{1}{c|}{0.1 decay}               & \multicolumn{1}{c|}{Random Crop}                              & \multirow{2}{*}{X}           \\
                   &                     &                     &      (every 20 epochs)                                     & \multicolumn{1}{c|}{Horizontal Flip}                            &                              \\ \cline{1-6}
\multicolumn{1}{|c|}{BAR}               & 256 x 256           & 64                  & \multicolumn{1}{c|}{0.1 decay}       & \multicolumn{1}{c|}{Random Crop}                                &          \multirow{2}{*}{O}                    \\
                   &                  &                     &         (every 20 epochs)                                    & \multicolumn{1}{c|}{Horizontal Flip}                            &                              \\ \cline{1-6}
\multicolumn{1}{|c|}{Dogs \& Cats}      & 224 x 224        & 64                  &    \multicolumn{1}{c|}{0.1 decay}                                      & \multicolumn{1}{c|}{Random Crop}          &  \multirow{2}{*}{O}                           \\
                   &                     &                     &         (every 20 epochs)                                    & \multicolumn{1}{c|}{Horizontal Flip}                            &                              \\ \hline
\end{tabular}%
}
\caption{Detailed training configurations for training our biased and debiased classifier, $f_B$ and $f_D$. The training configuration of image size, batch size, scheduler, augmentation and use of pretrained weights are mentioned for each dataset used.}
\label{model_configuration}
\end{table*}

\subsection{Further Analysis of Generated Samples}
We demonstrate examples of generated samples from the four benchmark datasets. As seen in Table~\ref{tab:performance}, our framework did not perform well in synthetic dataset. As our analysis in the main paper mentions that this may be due to the difficulty in capturing the semantics of the noises such as gaussian blur, frost and JPEG compression, the images in the first row of Table~\ref{tab:performance} clearly demonstrates that it is difficult to generate synthetic bias-conflict samples using our framework. Aside from CCIFAR10, we have successfully generated bias-conflict samples in the real-world datasets by capturing the semantics via image captioning method in our framework. Especially, in Dogs \& Cats dataset, our framework was able to generate ``bright fur cats'' and ``dark fur dogs'' which are expected bias-conflict pairs.




  